%% file: main.tex
\documentclass[10pt,twocolumn,letterpaper]{article}

\usepackage[pagenumbers]{wacv}  % arXiv version (camera-ready with page numbers)


\usepackage{multirow}

\definecolor{wacvblue}{rgb}{0.21,0.49,0.74}
\usepackage[pagebackref,breaklinks,colorlinks,allcolors=wacvblue]{hyperref}

\def\wacvPaperID{13}
\def\confName{WACV}
\def\confYear{2027}

\title{Satellite Pretraining Is Not One Thing: A Matched-Pair Study of Generic and\\Earth-Observation Foundation Models for Scene Classification}

\author{Sandarbh Agrawal \quad Rakshith Sathish \quad Ashwathi Nambiar\\
SatSure Analytics India Pvt.\ Ltd., Bengaluru, India\\
{\tt\small sandarbh15@gmail.com, rakshith@satsure.co, ashwathi@satsure.co}
}

\begin{document}
\maketitle
\input{sec/0_abstract}
\input{sec/1_intro}
\input{sec/2_related}
\input{sec/3_setup}
\input{sec/4_results}
\input{sec/5_analysis}
\input{sec/6_conclusion}
{
    \small
    \bibliographystyle{ieeenat_fullname}
    \bibliography{main}
}

\end{document}

%% file: sec/0_abstract.tex
\begin{abstract}
Geospatial foundation models are increasingly pretrained on satellite and aerial imagery, yet practitioners mapping
land use and land cover with few labels must still decide whether such an encoder is worth choosing over a strong
generic one. Prior comparisons rarely isolate the pretraining data: candidate encoders differ at once in objective,
architecture, size, data and preprocessing. We study the question with five \emph{matched pairs} --- an
Earth-observation (EO) encoder and a generic counterpart sharing objective and architecture (DINO/SatDINO,
DINOv3/DINOv3-SAT, MAE/SatMAE, MAE/Scale-MAE, CLIP/RemoteCLIP) --- embedded in a benchmark of 14 frozen encoders on
five RGB scene-classification datasets spanning 0.06--30\,m ground sample distance. Every encoder receives its native
preprocessing; we evaluate $k$-NN, linear and attentive probes as well as image retrieval over six random low-label splits,
with paired per-image significance tests. EO pretraining is not uniformly helpful: for the public checkpoints we study, on most datasets and metrics it
significantly improves three pairs (SatDINO, Scale-MAE and RemoteCLIP, the latter judged only on datasets it never
saw) and significantly hurts two (linear probe: DINOv3-SAT by up to 22.5 points, SatMAE by up to 8.8) --- including
opposite outcomes for SatMAE and
Scale-MAE, two masked autoencoders pretrained on the same fMoW-RGB imagery. The verdicts hold under patch-token
pooling and learned attentive probing, concentrate on small man-made objects such as courts, palaces and storage
tanks, and are largest when labels are scarcest (1--5\% of the data). Across encoders, the silhouette score of the
frozen embedding predicts $k$-NN accuracy and retrieval mAP almost perfectly (Spearman $\rho\geq0.96$ on four of five
datasets). For label-scarce EO scene classification, ``pretrained on satellite imagery'' is therefore not a reliable
selection criterion; each candidate checkpoint must be validated on the target data.
\end{abstract}

%% file: sec/1_intro.tex
\section{Introduction}
\label{sec:intro}

Mapping land use and land cover from overhead imagery underpins environmental monitoring, agriculture and urban
planning, but labelled data remain scarce and expensive to collect for every new region, sensor and class taxonomy.
Self-supervised foundation models~\cite{bommasani2021foundation} promise to close this gap: a frozen encoder
pretrained once on large unlabelled corpora is adapted to a new task with a handful of labels.
Two families compete for this role. \emph{Generic} encoders are pretrained on natural or web images ---
DINO~\cite{caron2021dino}, DINOv2/v3~\cite{oquab2024dinov2,simeoni2025dinov3}, MoCo~v3~\cite{chen2021mocov3},
MAE~\cite{he2022mae}, I-JEPA~\cite{assran2023ijepa} or CLIP~\cite{radford2021clip}. \emph{Earth-observation (EO)}
encoders are pretrained on satellite or aerial imagery~\cite{wang2022sslreview}, often by applying the same recipe to
a remote-sensing corpus: SatMAE~\cite{cong2022satmae} and Scale-MAE~\cite{reed2023scalemae} adapt masked
autoencoding, SatDINO~\cite{straka2025satdino} and DINOv3-SAT~\cite{simeoni2025dinov3} adapt self-distillation, and
RemoteCLIP~\cite{liu2024remoteclip} continues CLIP's image--text training on remote-sensing captions.

A practitioner choosing between them faces a simple question that is surprisingly hard to answer from the literature:
\emph{does EO pretraining actually buy anything over a strong generic encoder?} Earlier studies compared in-domain and
ImageNet pretraining for remote-sensing transfer with mixed conclusions~\cite{neumann2019indomain,risojevic2021imagenet}, and
Corley~\etal~\cite{corley2024revisiting} showed that once inputs are resized and normalised as during pretraining,
ImageNet models remain a strong baseline against many EO models. Benchmarks such as
GEO-Bench~\cite{lacoste2023geobench} and PANGAEA~\cite{marsocci2024pangaea} compare many EO foundation models across
tasks. Yet in such comparisons the candidates differ simultaneously in objective, architecture, model size,
pretraining data and preprocessing, so a ranking cannot tell \emph{why} one encoder beats another --- in particular,
whether the satellite pretraining data is responsible. Rankings are further confounded by evaluation details:
which token is used as the image feature, how features are normalised, and whether conclusions survive a probe that
can re-weight patch tokens~\cite{psomas2026ep}.

\begin{figure}[t]
  \centering
  \includegraphics[width=\linewidth]{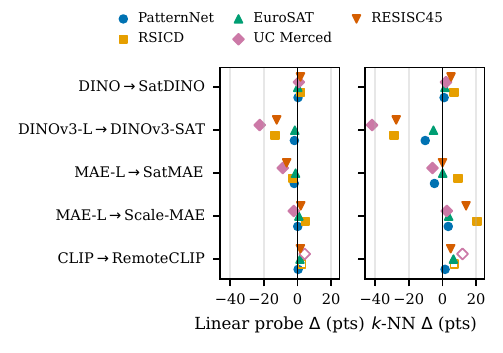}
  \caption{EO-pretrained minus generic counterpart for the five matched pairs on five datasets (percentage points,
  mean over six splits). Hollow markers: datasets RemoteCLIP saw during pretraining. With the
  linear probe (left), three EO encoders help and two hurt; $k$-NN (right) gives the same verdict for every pair except SatMAE, which is better on RSICD but worse on
  PatternNet and UC~Merced.}
  \label{fig:pairs}
\end{figure}

We address this with a controlled, \emph{matched-pair} design. For five EO encoders we evaluate a generic counterpart
with the same objective and architecture, so that the pretraining data (and, where unavoidable, details of the
recipe) is the main difference: DINO$\to$SatDINO, DINOv3$\to$DINOv3-SAT, MAE$\to$SatMAE, MAE$\to$Scale-MAE and
CLIP$\to$RemoteCLIP (\cref{tab:models}). The pairs sit inside a benchmark of 14 frozen encoders on five RGB
scene-classification datasets whose ground sample distance ranges from 0.06\,m to 30\,m. We focus on the low-label
regime that motivates foundation models in EO and on frozen features, which isolate representation quality from
fine-tuning choices. Our contributions are:
\begin{itemize}
\item A controlled comparison of EO and generic pretraining through five matched pairs, with native preprocessing for
every encoder, three probes plus retrieval, six random splits and paired per-image significance tests.
\item The finding that EO pretraining is not one thing: on most datasets it significantly helps three pairs and
significantly hurts two (\cref{fig:pairs}). SatMAE and Scale-MAE --- both masked autoencoders on the \emph{same} fMoW-RGB
imagery~\cite{christie2018fmow} --- land on opposite sides, so for these checkpoints the pretraining recipe matters as much as the domain.
\item Evidence that the verdicts are robust to the read-out (CLS, patch mean, attentive probing), concentrate on small
man-made objects, and grow as labels become scarce (1--5\%).
\item An analysis linking accuracy to the class separability of frozen embeddings, which explains both differences
between encoders and differences in difficulty between datasets.
\end{itemize}

%% file: sec/2_related.tex
\section{Related work}
\label{sec:related}

\paragraph{Self-supervised visual pretraining.}
Contrastive methods learn invariance between augmented views (SimCLR~\cite{chen2020simclr}, MoCo~\cite{he2020moco},
MoCo~v3~\cite{chen2021mocov3}); BYOL~\cite{grill2020byol} and DINO~\cite{caron2021dino} remove negatives through
self-distillation, a line continued by iBOT~\cite{zhou2022ibot}, DINOv2~\cite{oquab2024dinov2} and
DINOv3~\cite{simeoni2025dinov3}, which adds register tokens~\cite{darcet2024registers}. Masked image modelling
reconstructs hidden patches (BEiT~\cite{bao2022beit}, MAE~\cite{he2022mae}), and I-JEPA~\cite{assran2023ijepa} predicts
masked content in latent space. CLIP~\cite{radford2021clip} instead aligns images with captions. All encoders we evaluate
are Vision Transformers~\cite{dosovitskiy2021vit}, except for a supervised ResNet~\cite{he2016resnet} baseline
pretrained on ImageNet~\cite{deng2009imagenet}. A known property of masked models is that their features are
less linearly separable than those of joint-embedding methods~\cite{he2022mae,assran2023ijepa}, which makes the choice
of probe part of any comparison.

\paragraph{Earth-observation foundation models.}
Early EO self-supervision exploited geography and time: GASSL~\cite{ayush2021gassl} and SeCo~\cite{manas2021seco}
contrast views of the same location, and SSL4EO-S12~\cite{wang2023ssl4eo} provides a large multi-modal pretraining
corpus. SatMAE~\cite{cong2022satmae} and Scale-MAE~\cite{reed2023scalemae} adapt MAE to satellite imagery, the latter
with ground-sample-distance-aware positional encodings; CROMA~\cite{fuller2023croma} fuses radar and optical data;
Prithvi~\cite{jakubik2023prithvi}, SatlasPretrain~\cite{bastani2023satlas} and DOFA~\cite{xiong2024dofa} target
multi-spectral or multi-sensor inputs. SatDINO~\cite{straka2025satdino} trains DINO on fMoW~\cite{christie2018fmow},
DINOv3~\cite{simeoni2025dinov3} releases a satellite variant trained on 493M Maxar images, and
RemoteCLIP~\cite{liu2024remoteclip} continually pretrains CLIP on captioned remote-sensing data. Surveys review this
rapidly growing family~\cite{wang2022sslreview,bommasani2021foundation}.

\paragraph{Does in-domain pretraining help?}
Neumann~\etal~\cite{neumann2019indomain} and Risojevi\'c and Stojni\'c~\cite{risojevic2021imagenet} compared in-domain
and ImageNet pretraining for remote-sensing classification, and Corley~\etal~\cite{corley2024revisiting} showed that
matching the resize and normalisation used in pretraining makes ImageNet models a strong baseline against EO models.
GEO-Bench~\cite{lacoste2023geobench} and PANGAEA~\cite{marsocci2024pangaea} benchmark EO foundation models across
tasks and sensors. More broadly, ImageNet accuracy does not always predict transfer~\cite{kornblith2019transfer}, and
evaluation choices such as the architecture and the probe change the ranking of self-supervised
models~\cite{kolesnikov2019revisiting,psomas2026ep}. We build on
these works but differ in design: instead of ranking heterogeneous models, we compare \emph{matched} generic/EO pairs,
apply each model's native preprocessing~\cite{corley2024revisiting}, test robustness to the read-out with attentive
probing~\cite{psomas2026ep}, and report paired significance per split.

%% file: sec/3_setup.tex
\section{Experimental setup}
\label{sec:setup}

\input{tables/models}

\paragraph{Encoders and matched pairs.}
We evaluate 14 frozen encoders (\cref{tab:models}): a supervised ResNet-18~\cite{he2016resnet}; eight generic
self-supervised or image--text models (DINO~\cite{caron2021dino}, DINOv3 ViT-B and ViT-L~\cite{simeoni2025dinov3},
I-JEPA~\cite{assran2023ijepa}, MoCo~v3~\cite{chen2021mocov3}, MAE ViT-B and ViT-L~\cite{he2022mae},
CLIP~\cite{radford2021clip}); and five EO encoders (DINOv3-SAT~\cite{simeoni2025dinov3},
SatDINO~\cite{straka2025satdino}, SatMAE~\cite{cong2022satmae}, Scale-MAE~\cite{reed2023scalemae},
RemoteCLIP~\cite{liu2024remoteclip}). Each EO encoder is paired with the generic model that shares its objective and
architecture. The pairs control for the training objective and backbone, but not for every detail of the recipe: the
DINOv3 pair also differs in its pretraining corpus size (LVD-1689M vs.\ SAT-493M), Scale-MAE adds a scale-aware
positional encoding, and RemoteCLIP is initialised from the very CLIP checkpoint it is paired with (we measured a
mean cosine similarity of 0.999 between the two image encoders' weight matrices). SatDINO and Scale-MAE can take the
ground sample distance (GSD) as input; since our datasets do not provide a reliable per-image GSD, we drop SatDINO's
GSD token and use Scale-MAE's default resolution of 1\,m, as in the TorchGeo implementation~\cite{stewart2022torchgeo},
which may understate both models.

\begin{figure}[t]
  \centering
  \includegraphics[width=\linewidth]{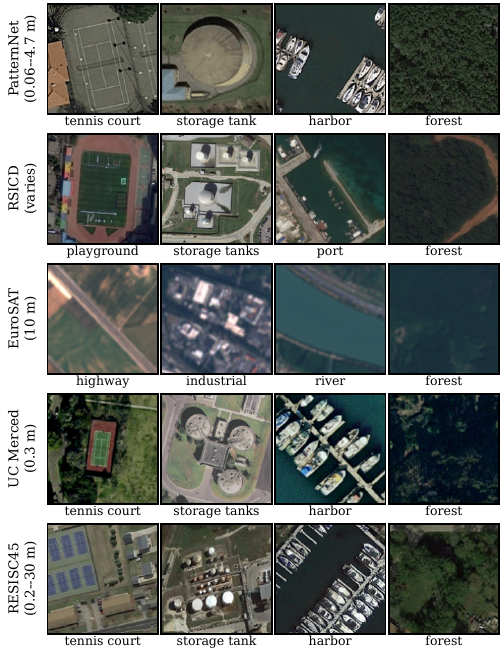}
  \caption{Example images from each dataset (four classes per dataset), with the dataset's ground sample distance. Three columns show small
  man-made objects --- the kind of class on which EO pretraining matters most in our per-class analysis
  (\cref{sec:results}); at EuroSAT's 10\,m resolution such objects are not resolvable, so its row shows land-cover
  classes.}
  \label{fig:samples}
\end{figure}

\paragraph{Datasets.}
We use five RGB scene-classification benchmarks with diverse sensors and resolutions (\cref{fig:samples}):
PatternNet~\cite{zhou2018patternnet} (30,400 aerial images, 38 classes, GSD 0.06--4.7\,m),
RSICD~\cite{lu2018rsicd} (8,734 images, 30 classes), EuroSAT~\cite{helber2019eurosat} (Sentinel-2 RGB, 10\,m GSD,
10 land-cover classes; we use the 16,200-image training split of the public release), UC~Merced~\cite{yang2010ucm}
(2,100 images, 21 classes, 0.3\,m) and RESISC45~\cite{cheng2017resisc} (31,500 images, 45 classes, GSD
0.2--30\,m). RSICD is a captioning dataset; we derive one label per image from the class name in the file name or,
failing that, from caption keywords. We restrict the study to RGB because it is the only input all 14 encoders accept;
a fair generic-vs-EO comparison is impossible on bands a generic model cannot read, and many EO encoders
(SatDINO, SatMAE's RGB variant, Scale-MAE, DINOv3-SAT, RemoteCLIP) are themselves RGB models.

\paragraph{Pretraining overlap.}
According to~\cite{liu2024remoteclip}, RemoteCLIP's pretraining data include the image--caption sets RSICD,
UCM-captions~\cite{qu2016ucmcaptions} (built on the UC~Merced images) and RSITMD~\cite{yuan2022rsitmd}.
RemoteCLIP has therefore seen two of our test sets; we report these numbers in grey and judge the CLIP$\to$RemoteCLIP
pair only on PatternNet, EuroSAT and RESISC45. The other EO encoders are pretrained on fMoW~\cite{christie2018fmow}
or Maxar imagery~\cite{simeoni2025dinov3}, which do not include our datasets. Web-scale corpora of the generic models
cannot be audited, a caveat shared by all such comparisons.

\paragraph{Preprocessing and features.}
Images are resized to 256 pixels (shorter side), centre-cropped to 224 and normalised with the statistics each
encoder was \emph{pretrained} with (\cref{tab:models}): ImageNet statistics for the generic models and Scale-MAE (as in
TorchGeo), CLIP statistics for CLIP and RemoteCLIP, and the SAT-493M and fMoW statistics released with DINOv3-SAT,
SatDINO and SatMAE~\cite{simeoni2025dinov3,straka2025satdino,cong2022satmae}, following the recommendation
of~\cite{corley2024revisiting}. The image-level feature is each encoder's own: the final-norm CLS token for the ViTs
(MAE's CLS token, used for linear probing in~\cite{he2022mae}), the average of patch tokens for I-JEPA, which has no
CLS token~\cite{assran2023ijepa}, and global average pooling for ResNet-18. For MAE we disable random masking at
inference (the Hugging Face implementation masks 75\% of patches by default). \Cref{sec:readout} repeats all
experiments with the mean of the final-layer patch tokens (register tokens~\cite{darcet2024registers} excluded) and
with attentive probing.

\paragraph{Probes and protocol.}
Each dataset is split into a stratified 10\% training and 90\% test set --- a low-label regime that stresses
representation quality --- and every experiment is repeated over six random splits (seeds 42, 1--5); we report the
mean (standard deviations are in the supplementary material). On the frozen embeddings we evaluate
(i) cosine $k$-NN ($k{=}20$; verdicts are unchanged for $k\in\{1,5,10\}$) on L2-normalised features;
(ii) a linear probe (LP): multinomial logistic regression (L-BFGS~\cite{liu1989lbfgs}, $C{=}1$) on features standardised
with training-split statistics, the scikit-learn~\cite{pedregosa2011sklearn} analogue of the parameter-free
BatchNorm that MAE places before its linear probe~\cite{he2022mae};
(iii) an MLP with one hidden layer of 256 units on the same standardised features (reported in the supplementary material);
(iv) retrieval mAP, ranking the training images for every test query by cosine similarity; and
(v) Efficient Probing (EP)~\cite{psomas2026ep}, an attentive probe with 32 learnable queries over the patch tokens,
trained with AdamW~\cite{loshchilov2019adamw} (1,000 steps, batch 128, weight decay $10^{-4}$, cosine schedule), its
learning rate chosen from $\{3{\cdot}10^{-4},10^{-3},3{\cdot}10^{-3}\}$ on a held-out 20\% of the seed-42 training split.
The LARS, batch-4096 recipe of~\cite{psomas2026ep} targets ImageNet and does not fit our 210--3,150 training images,
so we use the same fixed recipe for every encoder. A label-budget study repeats $k$-NN (all encoders, six splits) and the
linear probe (the nine pair members, three splits) at 1, 5, 10, 25 and 50\% training data. Models are run with
PyTorch~\cite{paszke2019pytorch}, timm~\cite{wightman2019timm}, Hugging Face Transformers~\cite{wolf2020transformers}
and OpenCLIP~\cite{ilharco2021openclip}; all encoders are public checkpoints, listed with every setting in the
supplementary material.

\paragraph{Statistical testing.}
The six splits share most of their test images, so a $t$-test across splits would overstate significance~\cite{nadeau2003inference}.
We therefore test each split separately and paired at the image level, as recommended
by~\cite{dietterich1998tests}: an exact McNemar test~\cite{mcnemar1947} on per-image correctness for the classifiers ($k$-NN, LP,
EP), and a Wilcoxon signed-rank test~\cite{wilcoxon1945} on per-query average precision for retrieval. Within each
split, $p$-values of all pair comparisons are corrected with the Holm procedure~\cite{holm1979}, and we report in how
many of the six splits a difference is significant at $\alpha{=}0.05$.

\paragraph{Separability.}
To relate accuracy to the geometry of the embedding we compute, per encoder and dataset, the silhouette
score~\cite{rousseeuw1987silhouettes} of the class labels under cosine distance (on a random 5,000-image sample) and
the ratio of mean within-class to mean between-centroid cosine distance, and we visualise embeddings with
t-SNE~\cite{vandermaaten2008tsne}.

%% file: tables/models.tex
\begin{table*}[t]
\centering
\footnotesize
\setlength{\tabcolsep}{4pt}
\begin{tabular}{@{}llllll@{}}
\toprule
Encoder & Params & Objective & Pretraining data & Feature & Normalisation\\
\midrule
\multicolumn{6}{l}{\textit{Supervised}}\\
ResNet-18 & 11M & supervised & ImageNet-1k (labels) & GAP & ImageNet\\
\midrule
\multicolumn{6}{l}{\textit{Generic (natural-image) pretraining}}\\
DINO ViT-B/16 & 86M & self-distillation & ImageNet-1k & CLS & ImageNet\\
DINOv3 ViT-B/16 & 86M & self-distillation$^\dagger$ & LVD-1689M (web) & CLS & ImageNet\\
DINOv3 ViT-L/16 & 303M & self-distillation$^\dagger$ & LVD-1689M (web) & CLS & ImageNet\\
I-JEPA ViT-H/14 & 631M & latent prediction & ImageNet-1k & patch mean & ImageNet\\
MoCo v3 ViT-B/16 & 86M & contrastive & ImageNet-1k & CLS & ImageNet\\
MAE ViT-B/16 & 86M & masked reconstruction & ImageNet-1k & CLS & ImageNet\\
MAE ViT-L/16 & 303M & masked reconstruction & ImageNet-1k & CLS & ImageNet\\
CLIP ViT-L/14 & 304M & image--text contrastive & WIT-400M (web) & CLS$^\ddagger$ & CLIP\\
\midrule
\multicolumn{6}{l}{\textit{EO-specific pretraining}}\\
DINOv3-SAT ViT-L/16 & 303M & self-distillation$^\dagger$ & SAT-493M (Maxar) & CLS & SAT-493M\\
SatDINO ViT-B/16 & 86M & self-distillation & fMoW-RGB & CLS & fMoW\\
SatMAE ViT-L/16 & 303M & masked reconstruction & fMoW-RGB & CLS & fMoW\\
Scale-MAE ViT-L/16 & 303M & masked recon.\ (scale-aware) & fMoW-RGB & CLS & ImageNet\\
RemoteCLIP ViT-L/14 & 304M & image--text contrastive & CLIP + RS captions & CLS$^\ddagger$ & CLIP\\
\bottomrule
\end{tabular}
\caption{The 14 frozen encoders. ``Feature'' is the image-level embedding used for $k$-NN, linear probe (LP)
and retrieval; ``Normalisation'' is the input mean/std each model was pretrained with, which we always apply.
$^\dagger$DINOv3 ViT-B/L and DINOv3-SAT ViT-L are distilled from 7B-parameter teachers~\cite{simeoni2025dinov3}.
$^\ddagger$CLIP's final-norm CLS token projected to the joint image--text space.
RemoteCLIP is OpenAI CLIP ViT-L/14 continually pretrained on remote-sensing image--caption pairs~\cite{liu2024remoteclip}.}
\label{tab:models}
\end{table*}

%% file: sec/4_results.tex
\section{Results}
\label{sec:results}

\input{tables/main}

\paragraph{Overall picture.}
\Cref{tab:main} reports $k$-NN, linear-probe and retrieval results for all 14 encoders. The strongest encoders come
from both families: averaged over the three datasets no encoder saw during pretraining and the three metrics, the
generic DINOv3 ViT-L ranks best (mean rank 2.4 of 14), followed by the EO models SatDINO (2.8) and RemoteCLIP (3.2);
RemoteCLIP has the best linear probe on RESISC45 (93.0 vs.\ 92.1 for DINOv3 ViT-L), the hardest unseen dataset, and
SatDINO the best on PatternNet and EuroSAT. The weakest encoders also come from both families: SatMAE (mean rank 12.8),
the generic MAE ViT-L (11.4) and DINOv3-SAT (11.1) rank at the bottom. Satellite pretraining thus neither
guarantees nor rules out a strong encoder.

\input{tables/pairs}

\paragraph{Matched pairs: EO pretraining helps three times and hurts twice.}
\Cref{tab:pairs,fig:pairs} isolate the effect of EO pretraining. Three EO encoders improve on their generic
counterparts. SatDINO beats DINO on every dataset and metric (linear probe $+0.1$ to $+1.9$ points, $k$-NN $+1.0$ to
$+6.8$, mAP $+3.8$ to $+11.7$), significantly in most cells. Scale-MAE beats MAE ViT-L in $k$-NN by $+2.6$ to
$+20.7$ points and in linear probe on four of five datasets, the exception being UC~Merced ($-2.2$, significant in only two of six
splits).
RemoteCLIP beats CLIP on all three datasets it did not see during pretraining, significantly for every metric
(linear probe $+0.4$ to $+1.7$, $k$-NN $+1.6$ to $+6.6$, mAP $+10.2$ to $+16.1$); its larger gains on RSICD and
UC~Merced are confounded by pretraining overlap (\cref{sec:setup}).
Two EO encoders are significantly \emph{worse} than their counterparts. DINOv3-SAT trails the web DINOv3 ViT-L ---
trained with the same recipe and distilled the same way --- on every dataset and metric in all six splits, by up to
22.5 linear-probe and 41.8 $k$-NN points on UC~Merced. SatMAE trails MAE ViT-L in linear probe on all five datasets
($-1.2$ to $-8.8$), while its $k$-NN and retrieval results are mixed. Its fMoW input statistics are not the cause: with
MAE's ImageNet statistics, SatMAE changes by about one point at most and still trails on all five datasets.

The two MAE pairs are the most informative contrast. SatMAE and Scale-MAE are both ViT-L masked autoencoders
pretrained on the same fMoW-RGB imagery~\cite{christie2018fmow}; they differ in the recipe --- most visibly
Scale-MAE's scale-aware positional encoding and its decoder that reconstructs low- and high-frequency images at
different scales~\cite{reed2023scalemae} --- yet Scale-MAE beats the ImageNet MAE in linear probe on four of the five
datasets while SatMAE trails it on all five. Likewise, DINOv3-SAT and
SatDINO both apply self-distillation to satellite imagery with opposite outcomes. For the public checkpoints we study, whether EO pretraining helps thus
depends on the specific model and its recipe, not on the domain of the pretraining data alone; with one checkpoint per
model, we cannot separate the contributions of the data and of the recipe.

\begin{figure*}[t]
  \centering
  \includegraphics[width=\linewidth]{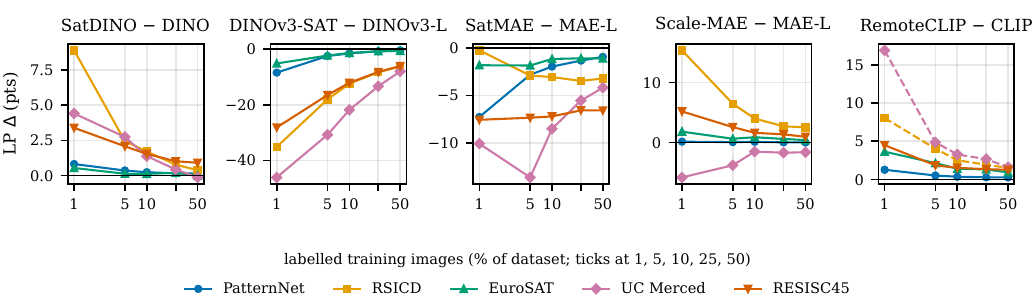}
  \caption{Label budget: EO-minus-generic linear-probe difference (points, mean over three splits) as the labelled
  training set grows from 1\% to 50\% of each dataset. Dashed: datasets RemoteCLIP saw during pretraining. At 1\%, UC~Merced has one labelled image per class. Gaps of
  either sign are largest when labels are scarcest. Per-encoder accuracies and $k$-NN results for all 14 encoders are
  in the supplementary material.}
  \label{fig:budget}
\end{figure*}

\paragraph{The gaps are largest when labels are scarce.}
\Cref{fig:budget} varies the labelled fraction from 1\% to 50\%. The direction of each pair's effect is largely stable across budgets --- it flips only where the gap is close to
zero, \eg SatDINO on EuroSAT and UC~Merced at 50\% --- but its magnitude shrinks as labels grow. With 1\% of the labels, SatDINO leads DINO by 8.9 points on RSICD and 4.4 on
UC~Merced, Scale-MAE leads MAE ViT-L by 15.4 points on RSICD, and RemoteCLIP leads CLIP by 4.5 points on RESISC45 ---
gaps that fall to 0.4, $-0.1$, 2.5 and 1.3 points at 50\%. Conversely, DINOv3-SAT trails DINOv3 ViT-L by 46.1 linear-probe points
on UC~Merced and 28.2 on RESISC45 at 1\%, versus 8.0 and 6.1 at 50\%. The choice of encoder therefore matters most in
the label-scarce regime that motivates foundation models for EO mapping, while with abundant labels the probe can
partly compensate for a poorer representation.

\input{tables/per_class}

\paragraph{Where EO pretraining helps or hurts.}
Per-class linear-probe accuracy on RESISC45 (\cref{tab:perclass}) shows that the effect of EO pretraining is
concentrated on classes defined by \emph{small man-made objects}. The two EO models that hurt fail precisely there:
DINOv3-SAT loses 34.7, 34.4 and 33.0 points on basketball courts, palaces and tennis courts, and SatMAE 19.9, 19.5 and
17.4 points on palaces, airports and storage tanks. The EO models that help gain on the same kind of classes:
RemoteCLIP on tennis courts ($+13.0$), basketball courts and baseball diamonds; Scale-MAE on baseball diamonds
($+8.5$) and ground track fields; SatDINO on railways ($+7.0$) and tennis courts. Natural scene classes change much less: over
thirteen of them (\eg forest, desert, lake, wetland, sea ice) the mean absolute per-class change is 0.8--3.0 points
per pair, against 2.1--12.7 points over all 45 classes. In other words, the pretraining data and recipe matter most for
recognising small structures whose apparent size depends on the ground sample distance --- an observation consistent
with the different native resolutions of the pretraining corpora (\eg 0.6\,m Maxar imagery for
DINOv3-SAT~\cite{simeoni2025dinov3}) and with Scale-MAE's explicit modelling of scale, although our experiments do not
establish this mechanism causally.

%% file: tables/main.tex
\begin{table*}[t]
\centering
\footnotesize
\setlength{\tabcolsep}{2.9pt}
\begin{tabular}{@{}lccccccccccccccc@{}}
\toprule
 & \multicolumn{3}{c}{PatternNet} & \multicolumn{3}{c}{RSICD} & \multicolumn{3}{c}{EuroSAT} & \multicolumn{3}{c}{UC Merced} & \multicolumn{3}{c}{RESISC45}\\
\cmidrule(lr){2-4}\cmidrule(lr){5-7}\cmidrule(lr){8-10}\cmidrule(lr){11-13}\cmidrule(lr){14-16}
Encoder & $k$-NN & LP & mAP & $k$-NN & LP & mAP & $k$-NN & LP & mAP & $k$-NN & LP & mAP & $k$-NN & LP & mAP\\
\midrule
ResNet-18 & 92.9 & 96.6 & 62.7 & 70.0 & 76.7 & 35.1 & 86.9 & 90.9 & 56.8 & 73.5 & 85.0 & 57.6 & 71.6 & 77.7 & 29.8\\
\midrule
DINO ViT-B/16 & 95.3 & \underline{98.8} & 72.4 & 76.5 & 87.2 & 43.1 & 90.7 & \underline{96.2} & 62.8 & 78.4 & 92.7 & 64.5 & 78.7 & 89.0 & 38.1\\
DINOv3 ViT-B/16 & 96.3 & 98.6 & 78.1 & 82.8 & 89.7 & \underline{58.5} & 89.7 & 95.4 & 60.1 & \underline{82.3} & \underline{94.3} & \underline{77.3} & 83.1 & 89.9 & 47.9\\
DINOv3 ViT-L/16 & \underline{96.8} & 98.6 & \textbf{84.4} & \textbf{84.5} & \textbf{91.1} & \textbf{64.3} & \underline{92.0} & 95.9 & \underline{63.7} & \textbf{84.9} & \textbf{95.2} & \textbf{82.9} & \underline{87.0} & \underline{92.1} & \underline{58.6}\\
I-JEPA ViT-H/14 & 84.8 & 97.8 & 47.0 & 60.2 & 82.4 & 27.0 & 86.4 & 95.2 & 54.4 & 53.5 & 85.0 & 40.4 & 60.1 & 83.5 & 21.7\\
MoCo v3 ViT-B/16 & 93.5 & 98.6 & 63.8 & 72.6 & 85.4 & 37.6 & 89.8 & 95.6 & 58.9 & 73.0 & 90.9 & 57.8 & 74.7 & 87.1 & 32.7\\
MAE ViT-B/16 & 91.5 & 97.9 & 59.4 & 65.4 & 81.3 & 28.9 & 84.9 & 93.0 & 52.8 & 56.4 & 83.6 & 41.9 & 65.2 & 81.7 & 23.1\\
MAE ViT-L/16 & 88.1 & 98.3 & 49.9 & 55.4 & 83.7 & 23.4 & 83.5 & 94.9 & 46.9 & 47.5 & 87.2 & 34.9 & 59.3 & 84.3 & 19.9\\
CLIP ViT-L/14 & 95.5 & 98.5 & 73.4 & \underline{84.4} & \underline{91.0} & 58.1 & 81.4 & 93.5 & 43.8 & 76.9 & 92.4 & 64.1 & 85.4 & 91.3 & 51.0\\
\midrule
DINOv3-SAT ViT-L/16 & 86.6 & 96.7 & 53.1 & 55.8 & 77.8 & 26.3 & 86.6 & 94.3 & 52.5 & 43.2 & 72.7 & 35.3 & 59.6 & 79.7 & 24.9\\
SatDINO ViT-B/16 & 96.3 & \textbf{99.1} & 77.2 & 83.3 & 89.1 & 54.8 & \textbf{92.2} & \textbf{96.4} & \textbf{67.3} & 80.5 & 93.5 & 68.3 & 83.9 & 90.7 & 44.3\\
SatMAE ViT-L/16 & 83.4 & 96.3 & 45.5 & 64.6 & 81.1 & 29.1 & 83.7 & 93.7 & 53.3 & 41.6 & 78.4 & 31.4 & 59.3 & 77.8 & 19.2\\
Scale-MAE ViT-L/16 & 91.6 & 98.4 & 58.3 & 76.0 & 88.4 & 40.0 & 87.2 & 95.7 & 53.8 & 50.1 & 85.0 & 36.6 & 73.3 & 86.2 & 28.6\\
RemoteCLIP ViT-L/14$^{*}$ & \textbf{97.1} & \underline{98.8} & \underline{83.7} & \textcolor{gray}{91.4} & \textcolor{gray}{93.4} & \textcolor{gray}{72.3} & 88.0 & 94.9 & 54.9 & \textcolor{gray}{88.8} & \textcolor{gray}{96.6} & \textcolor{gray}{81.5} & \textbf{90.3} & \textbf{93.0} & \textbf{67.1}\\
\bottomrule
\end{tabular}
\caption{Frozen-feature results (\%, mean over six random 10\%/90\% train/test splits; standard deviations in the
supplementary material). \textbf{Bold}/\underline{underline}: best/second-best per column.
$^{*}$RemoteCLIP was pretrained on RSICD and UC~Merced image--caption data~\cite{liu2024remoteclip}; its
\textcolor{gray}{grey} entries are not comparable and are excluded from the ranking.}
\label{tab:main}
\end{table*}

%% file: tables/pairs.tex
\begin{table*}[t]
\centering
\footnotesize
\setlength{\tabcolsep}{2.6pt}
\begin{tabular}{@{}lccccccccccccccc@{}}
\toprule
EO model & \multicolumn{3}{c}{PatternNet} & \multicolumn{3}{c}{RSICD} & \multicolumn{3}{c}{EuroSAT} & \multicolumn{3}{c}{UC Merced} & \multicolumn{3}{c}{RESISC45}\\
\cmidrule(lr){2-4}\cmidrule(lr){5-7}\cmidrule(lr){8-10}\cmidrule(lr){11-13}\cmidrule(lr){14-16}
 & $k$-NN & LP & mAP & $k$-NN & LP & mAP & $k$-NN & LP & mAP & $k$-NN & LP & mAP & $k$-NN & LP & mAP\\
\midrule
SatDINO & +1.0$^{\blacktriangle}$ & +0.3$^{\blacktriangle}$ & +4.8$^{\blacktriangle}$ & +6.8$^{\blacktriangle}$ & +1.9$^{\blacktriangle}$ & +11.7$^{\blacktriangle}$ & +1.5$^{\blacktriangle}$ & +0.1 & +4.4$^{\blacktriangle}$ & +2.1 & +0.8 & +3.8$^{\blacktriangle}$ & +5.2$^{\blacktriangle}$ & +1.7$^{\blacktriangle}$ & +6.2$^{\blacktriangle}$\\
DINOv3-SAT & $-$10.2$^{\triangledown}$ & $-$1.9$^{\triangledown}$ & $-$31.3$^{\triangledown}$ & $-$28.7$^{\triangledown}$ & $-$13.3$^{\triangledown}$ & $-$37.9$^{\triangledown}$ & $-$5.3$^{\triangledown}$ & $-$1.6$^{\triangledown}$ & $-$11.2$^{\triangledown}$ & $-$41.8$^{\triangledown}$ & $-$22.5$^{\triangledown}$ & $-$47.6$^{\triangledown}$ & $-$27.4$^{\triangledown}$ & $-$12.4$^{\triangledown}$ & $-$33.7$^{\triangledown}$\\
SatMAE & $-$4.7$^{\triangledown}$ & $-$2.0$^{\triangledown}$ & $-$4.4$^{\triangledown}$ & +9.3$^{\blacktriangle}$ & $-$2.6$^{\triangledown}$ & +5.7$^{\blacktriangle}$ & +0.2 & $-$1.2$^{\triangledown}$ & +6.4$^{\blacktriangle}$ & $-$5.9$^{\triangledown}$ & $-$8.8$^{\triangledown}$ & $-$3.5$^{\triangledown}$ & +0.0 & $-$6.5$^{\triangledown}$ & $-$0.8$^{\triangledown}$\\
Scale-MAE & +3.5$^{\blacktriangle}$ & +0.1 & +8.3$^{\blacktriangle}$ & +20.7$^{\blacktriangle}$ & +4.7$^{\blacktriangle}$ & +16.7$^{\blacktriangle}$ & +3.7$^{\blacktriangle}$ & +0.8$^{\blacktriangle}$ & +6.8$^{\blacktriangle}$ & +2.6 & $-$2.2 & +1.7$^{\blacktriangle}$ & +14.0$^{\blacktriangle}$ & +1.9$^{\blacktriangle}$ & +8.7$^{\blacktriangle}$\\
RemoteCLIP & +1.6$^{\blacktriangle}$ & +0.4$^{\blacktriangle}$ & +10.2$^{\blacktriangle}$ & \textcolor{gray}{+7.1$^{\blacktriangle}$} & \textcolor{gray}{+2.5$^{\blacktriangle}$} & \textcolor{gray}{+14.2$^{\blacktriangle}$} & +6.6$^{\blacktriangle}$ & +1.5$^{\blacktriangle}$ & +11.1$^{\blacktriangle}$ & \textcolor{gray}{+11.9$^{\blacktriangle}$} & \textcolor{gray}{+4.2$^{\blacktriangle}$} & \textcolor{gray}{+17.4$^{\blacktriangle}$} & +4.9$^{\blacktriangle}$ & +1.7$^{\blacktriangle}$ & +16.1$^{\blacktriangle}$\\
\bottomrule
\end{tabular}
\caption{Matched pairs: each EO model minus its generic counterpart (SatDINO--DINO, DINOv3-SAT--DINOv3 ViT-L,
SatMAE--MAE ViT-L, Scale-MAE--MAE ViT-L, RemoteCLIP--CLIP; percentage points, mean over six splits).
$^{\blacktriangle}$/$^{\triangledown}$: the EO / generic model is significantly better in at least 5 of the 6
splits (exact McNemar test for LP and $k$-NN, Wilcoxon signed-rank test on per-query AP for mAP; Holm-corrected
within each split). \textcolor{gray}{Grey}: datasets RemoteCLIP saw during pretraining.}
\label{tab:pairs}
\end{table*}

%% file: tables/per_class.tex
\begin{table*}[t]
\centering
\footnotesize
\setlength{\tabcolsep}{4pt}
\begin{tabular}{@{}lp{0.385\linewidth}p{0.385\linewidth}@{}}
\toprule
EO model & Largest per-class gains of the EO model & Largest per-class losses of the EO model\\
\midrule
SatDINO & railway (+7.0), tennis court (+5.7), basketball court (+5.1) & river ($-$1.8), roundabout ($-$1.3), cloud ($-$1.2)\\
DINOv3-SAT & mountain (+5.9), chaparral (+0.5), desert ($-$0.4) & basketball court ($-$34.7), palace ($-$34.4), tennis court ($-$33.0)\\
SatMAE & freeway (+3.7), intersection (+0.1), sea ice (+0.0) & palace ($-$19.9), airport ($-$19.5), storage tank ($-$17.4)\\
Scale-MAE & baseball diamond (+8.5), ground track field (+7.7), freeway (+7.3) & rectangular farmland ($-$3.4), lake ($-$1.9), island ($-$1.8)\\
RemoteCLIP & tennis court (+13.0), basketball court (+7.7), baseball diamond (+6.6) & wetland ($-$2.7), commercial area ($-$2.6), palace ($-$1.6)\\
\bottomrule
\end{tabular}
\caption{RESISC45 classes with the largest change in linear-probe per-class accuracy between each EO model and its
generic counterpart (as in \cref{tab:pairs}; points, mean
over six splits). Values are signed: a negative entry among the ``gains'' means the EO model improves on fewer
than three classes.}
\label{tab:perclass}
\end{table*}

%% file: sec/5_analysis.tex
\section{Analysis}
\label{sec:analysis}

\subsection{Are the verdicts artefacts of the read-out?}
\label{sec:readout}

\input{tables/robustness}

Frozen-feature comparisons are sensitive to how an image-level feature is formed: a CLS token can discard
information that a model stores in its patch tokens, which particularly concerns masked
models~\cite{he2022mae,assran2023ijepa,psomas2026ep}. We therefore re-ran every experiment with the mean of the
final-layer patch tokens for all ViTs, and trained an attentive Efficient Probe (EP)~\cite{psomas2026ep} on the patch
tokens of every ViT, which learns its own aggregation.

\Cref{tab:robustness} shows that four of the five pair verdicts are invariant to the read-out: SatDINO, Scale-MAE and
RemoteCLIP stay ahead, and DINOv3-SAT stays behind, under native and patch-mean $k$-NN, both linear probes and EP, on
almost all datasets. The one consistent exception is Scale-MAE on UC~Merced, where it trails MAE ViT-L under all
three trained probes (EP: $-3.8$ points, significant in all six splits). The paired tests agree: with EP, SatDINO
remains significantly better than DINO on PatternNet, RSICD and RESISC45, RemoteCLIP than CLIP on EuroSAT and
RESISC45, and DINOv3-SAT remains significantly worse than DINOv3 on four datasets, tying only on the near-saturated
PatternNet. The fifth pair, SatMAE, is read-out-dependent for $k$-NN only: its mean $k$-NN difference
flips from $-0.2$ to $+12.0$ points under patch-mean pooling, mainly because MAE ViT-L's own patch-mean $k$-NN
collapses. For trained probes SatMAE stays behind under every read-out.

Two broader regularities emerge (supplementary material). First, the trained linear probe is far more forgiving than
$k$-NN for masked and predictive models: LP exceeds $k$-NN by 22.9, 19.8 and 19.0 points on average for MAE ViT-L,
I-JEPA and SatMAE, but only by 4.2 and 5.5 for RemoteCLIP and DINOv3 ViT-L, so rankings based on $k$-NN or retrieval
penalise masked models much more. Second, EP changes the linear-probe accuracy by less than one point on average for
every encoder except DINOv3-SAT ($+6.0$). With 210--3,150 training images, a learned attention pool brings no
systematic benefit over a linear probe on a well-chosen token, so the pair verdicts are not an artefact of the linear probe.

\input{tables/dinov3_ablation}

\paragraph{Why does DINOv3-SAT fall so far behind?}
DINOv3-SAT's gap to its web counterpart is the largest in our study. It is consistent with the DINOv3 authors' own
frozen RGB evaluation on GEO-Bench, where the web 7B model beats the satellite 7B model on m-eurosat (97.0 vs.\ 94.8)
and on five of six classification tasks, while the satellite model leads in canopy-height
estimation~\cite{simeoni2025dinov3}; the web ViT-L is not evaluated there, which our pair adds. We nevertheless checked
whether the gap stems from our evaluation (\cref{tab:dinov3}). Its CLS token is indeed a poor scene descriptor: averaging its patch tokens raises UC~Merced
$k$-NN from 43.2 to 60.5 and linear probe from 72.7 to 84.9, whereas the web model is almost insensitive to the choice
of feature. Its native SAT-493M normalisation helps only slightly over ImageNet statistics (patch mean: 60.5 vs.\
56.7 $k$-NN). Even with its best feature, however, DINOv3-SAT trails the web model by 25.4 $k$-NN and 10.6 linear-probe
points on UC~Merced and by 10.3 and 6.2 on RSICD. The deficit is thus real for frozen scene classification, and
consistent with a model whose information resides in patch tokens tuned to 0.6\,m Maxar imagery and dense
prediction~\cite{simeoni2025dinov3}.

\subsection{Class separability tracks neighbourhood-based accuracy}
\label{sec:separability}

\begin{figure}[t]
  \centering
  \includegraphics[width=\linewidth]{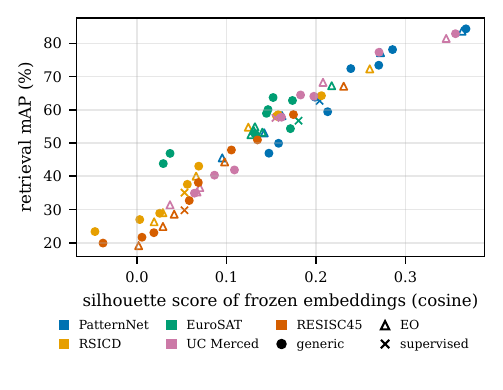}
  \caption{Silhouette score of the frozen embeddings vs.\ retrieval mAP for all 14 encoders on five datasets
  (70 points). Spearman $\rho=0.97$ across all points; within each dataset $\rho\geq0.85$.}
  \label{fig:sep}
\end{figure}

Why do some encoders transfer better, and why are some datasets harder? We relate accuracy to the silhouette
score~\cite{rousseeuw1987silhouettes} of the frozen embeddings, which measures how much closer an image is to its own
class than to the nearest other class. Across the 14 encoders, silhouette predicts accuracy almost perfectly within
each dataset: Spearman $\rho$ with $k$-NN accuracy is 0.96--0.98 on PatternNet, RSICD, UC~Merced and RESISC45, and
$\rho$ with retrieval mAP is 0.97--0.996 on the same datasets (\cref{fig:sep}). The exception is EuroSAT ($\rho=0.68$
for $k$-NN, 0.86 for mAP), presumably because its ten coarse land-cover classes are separated well enough by most
encoders that the remaining differences in accuracy no longer track geometry closely. The ratio of within-class to between-class distance gives a similar picture with
opposite sign ($\rho$ from $-0.74$ to $-0.96$ with $k$-NN) on the same four datasets, but not on EuroSAT. The agreement
is expected, as $k$-NN, retrieval and the silhouette all rely on cosine neighbourhoods; its practical value is that the
silhouette needs no trained probe and only a small labelled sample. Computed on each split's labelled 10\% training
set alone (210--3,150 images), it ranks the encoders as the 5,000-image estimate does ($\rho\geq0.97$) and gives the same
correlations ($\rho=0.96$--$0.98$ with $k$-NN, 0.97--0.996 with mAP, on the four datasets). On RESISC45,
RemoteCLIP raises CLIP's silhouette from 0.13 to 0.23, while DINOv3-SAT lowers DINOv3's from 0.17 to 0.03 (t-SNE
projections in the supplementary material).

Separability also accounts for most of the difference in difficulty between datasets, which a summary by class count
or resolution does not: PatternNet, with the most classes (38) but the highest mean silhouette (0.22), is the easiest,
and RESISC45 and RSICD, with the lowest (0.07 and 0.08), are the hardest. UC~Merced is the outlier --- intermediate
separability (between EuroSAT and PatternNet) but the lowest mean $k$-NN accuracy (66.5\%) --- because its 10\% split leaves only ten training
images per class, which penalises $k$-NN with $k{=}20$ in particular.

\paragraph{Model size helps only some recipes.}
Within one recipe, more parameters are not always better. Going from DINOv3 ViT-B to ViT-L improves $k$-NN and mAP on every
dataset ($k$-NN $+0.5$ to $+3.9$, mAP $+3.6$ to $+10.7$ points), whereas going from MAE ViT-B to ViT-L lowers $k$-NN by 1.4 to
10.1 points and mAP by 3.1 to 9.5 while raising the linear probe by 0.4 to 3.6: the larger masked model's features are
more linearly separable but less well organised for nearest-neighbour use.

%% file: tables/robustness.tex
\begin{table}[t]
\centering
\scriptsize
\setlength{\tabcolsep}{2.4pt}
\begin{tabular}{@{}lccccc@{}}
\toprule
 & \multicolumn{2}{c}{$k$-NN} & \multicolumn{3}{c}{Trained probe}\\
\cmidrule(lr){2-3}\cmidrule(lr){4-6}
EO model & native & patch & LP & LP patch & EP\\
\midrule
SatDINO & +3.3 (5/5) & +2.6 (5/5) & +1.0 (5/5) & +1.1 (5/5) & +1.1 (5/5)\\
DINOv3-SAT & $-$22.7 (0/5) & $-$11.7 (0/5) & $-$10.3 (0/5) & $-$5.0 (0/5) & $-$4.6 (0/5)\\
SatMAE & $-$0.2 (3/5) & +12.0 (5/5) & $-$4.2 (0/5) & $-$2.6 (0/5) & $-$2.9 (0/5)\\
Scale-MAE & +8.9 (5/5) & +11.4 (5/5) & +1.0 (4/5) & +0.8 (4/5) & +0.8 (4/5)\\
RemoteCLIP & +4.4 (3/3) & +2.8 (3/3) & +1.2 (3/3) & +0.8 (3/3) & +0.9 (3/3)\\
\bottomrule
\end{tabular}
\caption{Does the verdict depend on how features are read out? Mean difference between each EO model and its
generic counterpart (as in \cref{tab:pairs}; points) over
the comparable datasets (five; three for RemoteCLIP), and in parentheses the number of datasets where the EO model is
ahead. ``native'': the encoder's own feature (\cref{tab:models}); ``patch'': mean of final-layer patch tokens;
EP: attentive probe on patch tokens~\cite{psomas2026ep}.}
\label{tab:robustness}
\end{table}

%% file: tables/dinov3_ablation.tex
\begin{table}[t]
\centering
\footnotesize
\setlength{\tabcolsep}{2.4pt}
\begin{tabular}{@{}llcccccc@{}}
\toprule
 & & \multicolumn{3}{c}{UC Merced} & \multicolumn{3}{c}{RSICD}\\
\cmidrule(lr){3-5}\cmidrule(lr){6-8}
DINOv3-L & Feature & $k$-NN & LP & mAP & $k$-NN & LP & mAP\\
\midrule
web & CLS & 84.9 & 95.2 & 82.9 & 84.5 & 91.1 & 64.3\\
web & patch mean & 85.9 & 95.2 & 84.4 & 86.4 & 91.4 & 68.3\\
web & CLS $\oplus$ patch & 85.6 & 95.6 & 83.9 & 85.4 & 91.8 & 66.0\\
\midrule
SAT & CLS & 43.2 & 72.7 & 35.3 & 55.8 & 77.8 & 26.3\\
SAT & patch mean & 60.5 & 84.9 & 51.7 & 76.1 & 85.6 & 47.5\\
SAT & CLS $\oplus$ patch & 44.7 & 81.9 & 37.0 & 59.2 & 84.9 & 28.6\\
\midrule
SAT (IN norm.) & CLS & 45.6 & 74.8 & 36.5 & 57.5 & 78.8 & 26.9\\
SAT (IN norm.) & patch mean & 56.7 & 84.2 & 48.6 & 73.5 & 84.9 & 44.8\\
SAT (IN norm.) & CLS $\oplus$ patch & 47.3 & 82.0 & 37.9 & 60.2 & 84.5 & 29.0\\
\bottomrule
\end{tabular}
\caption{DINOv3 ViT-L web vs.\ SAT (six splits): image-level feature and input normalisation. ``IN norm.'':
SAT model fed ImageNet statistics instead of its own SAT-493M statistics.}
\label{tab:dinov3}
\end{table}

%% file: sec/6_conclusion.tex
\section{Discussion and conclusion}
\label{sec:conclusion}

\paragraph{Implications for practitioners.}
For label-scarce land-use and land-cover scene classification from RGB imagery, the label ``pretrained on satellite
imagery'' is not a reliable criterion for choosing a frozen encoder. In our matched comparisons, EO pretraining
improved three encoders and degraded two, with differences of up to 46 linear-probe points at 1\% labels. We therefore recommend
(i) validating candidate encoders on a small labelled sample of the target data, where the differences are largest;
(ii) feeding each encoder its own pretraining normalisation and checking more than one read-out --- a $k$-NN or
retrieval comparison alone strongly penalises masked models, while for some EO models (DINOv3-SAT) patch tokens carry
far more information than the CLS token; and
(iii) using the silhouette score of the frozen embeddings on that sample as a cheap, label-efficient predictor of
$k$-NN and retrieval quality.

\paragraph{Limitations.}
Our conclusions are bounded by the design. (1)~We study RGB scene classification with frozen encoders; EO
pretraining may pay off differently for multi-spectral or SAR inputs~\cite{fuller2023croma,xiong2024dofa,wang2023ssl4eo},
for dense tasks such as segmentation~\cite{lacoste2023geobench,marsocci2024pangaea,simeoni2025dinov3}, or after
fine-tuning. (2)~The pairs control objective and architecture but not every factor: corpus size and content,
pretraining length and recipe details differ, so our results show \emph{that} EO pretraining can help or hurt, and on
which classes, but not causally \emph{which} ingredient is responsible; the small-object pattern suggests a role for
the resolution of the pretraining imagery that future work could test with controlled rescaling. (3)~SatDINO and
Scale-MAE are run without their GSD input, which may understate them. (4)~RemoteCLIP saw two of our test sets during
pretraining; we exclude them from its comparison, but generic web-scale corpora cannot be audited for overlap.
(5)~RSICD labels are derived from file names and captions and are noisier than the other datasets' labels, and all
datasets are moderate in size.

\paragraph{Conclusion.}
Five matched pairs show that satellite pretraining is not one thing: depending on the recipe it makes an encoder
clearly better or clearly worse than its generic counterpart, most visibly on small man-made structures and with
scarce labels, largely regardless of the read-out. We hope such matched comparisons will complement leaderboard benchmarks of
geospatial foundation models.

\paragraph{Acknowledgements.}
This work was supported by SatSure Analytics India Pvt.\ Ltd., which provided the computational resources.

%% file: main.bbl
\begin{thebibliography}{60}
\providecommand{\natexlab}[1]{#1}
\providecommand{\url}[1]{\texttt{#1}}
\expandafter\ifx\csname urlstyle\endcsname\relax
  \providecommand{\doi}[1]{doi: #1}\else
  \providecommand{\doi}{doi: \begingroup \urlstyle{rm}\Url}\fi

\bibitem[Assran et~al.(2023)Assran, Duval, Misra, Bojanowski, Vincent, Rabbat,
  LeCun, and Ballas]{assran2023ijepa}
Mahmoud Assran, Quentin Duval, Ishan Misra, Piotr Bojanowski, Pascal Vincent,
  Michael Rabbat, Yann LeCun, and Nicolas Ballas.
\newblock Self-supervised learning from images with a joint-embedding
  predictive architecture.
\newblock In \emph{IEEE/CVF Conference on Computer Vision and Pattern
  Recognition (CVPR)}, pages 15619--15629, 2023.

\bibitem[Ayush et~al.(2021)Ayush, Uzkent, Meng, Tanmay, Burke, Lobell, and
  Ermon]{ayush2021gassl}
Kumar Ayush, Burak Uzkent, Chenlin Meng, Kumar Tanmay, Marshall Burke, David
  Lobell, and Stefano Ermon.
\newblock Geography-aware self-supervised learning.
\newblock In \emph{IEEE/CVF International Conference on Computer Vision
  (ICCV)}, pages 10161--10170, 2021.

\bibitem[Bao et~al.(2022)Bao, Dong, Piao, and Wei]{bao2022beit}
Hangbo Bao, Li Dong, Songhao Piao, and Furu Wei.
\newblock {BEiT}: {BERT} pre-training of image transformers.
\newblock In \emph{International Conference on Learning Representations
  (ICLR)}, 2022.

\bibitem[Bastani et~al.(2023)Bastani, Wolters, Gupta, Ferdinando, and
  Kembhavi]{bastani2023satlas}
Favyen Bastani, Piper Wolters, Ritwik Gupta, Joe Ferdinando, and Aniruddha
  Kembhavi.
\newblock {SatlasPretrain}: A large-scale dataset for remote sensing image
  understanding.
\newblock In \emph{IEEE/CVF International Conference on Computer Vision
  (ICCV)}, pages 16726--16736, 2023.

\bibitem[Bommasani et~al.(2021)Bommasani, Hudson, Adeli, Altman, Arora,
  et~al.]{bommasani2021foundation}
Rishi Bommasani, Drew~A. Hudson, Ehsan Adeli, Russ Altman, Simran Arora, et~al.
\newblock On the opportunities and risks of foundation models.
\newblock \emph{arXiv preprint arXiv:2108.07258}, 2021.

\bibitem[Caron et~al.(2021)Caron, Touvron, Misra, J{\'e}gou, Mairal,
  Bojanowski, and Joulin]{caron2021dino}
Mathilde Caron, Hugo Touvron, Ishan Misra, Herv{\'e} J{\'e}gou, Julien Mairal,
  Piotr Bojanowski, and Armand Joulin.
\newblock Emerging properties in self-supervised vision transformers.
\newblock In \emph{IEEE/CVF International Conference on Computer Vision
  (ICCV)}, pages 9630--9640, 2021.

\bibitem[Chen et~al.(2020)Chen, Kornblith, Norouzi, and Hinton]{chen2020simclr}
Ting Chen, Simon Kornblith, Mohammad Norouzi, and Geoffrey Hinton.
\newblock A simple framework for contrastive learning of visual
  representations.
\newblock In \emph{International Conference on Machine Learning (ICML)}, pages
  1597--1607, 2020.

\bibitem[Chen et~al.(2021)Chen, Xie, and He]{chen2021mocov3}
Xinlei Chen, Saining Xie, and Kaiming He.
\newblock An empirical study of training self-supervised vision transformers.
\newblock In \emph{IEEE/CVF International Conference on Computer Vision
  (ICCV)}, pages 9620--9629, 2021.

\bibitem[Cheng et~al.(2017)Cheng, Han, and Lu]{cheng2017resisc}
Gong Cheng, Junwei Han, and Xiaoqiang Lu.
\newblock Remote sensing image scene classification: Benchmark and state of the
  art.
\newblock \emph{Proceedings of the IEEE}, 105\penalty0 (10):\penalty0
  1865--1883, 2017.

\bibitem[Christie et~al.(2018)Christie, Fendley, Wilson, and
  Mukherjee]{christie2018fmow}
Gordon Christie, Neil Fendley, James Wilson, and Ryan Mukherjee.
\newblock Functional map of the world.
\newblock In \emph{IEEE/CVF Conference on Computer Vision and Pattern
  Recognition (CVPR)}, pages 6172--6180, 2018.

\bibitem[Cong et~al.(2022)Cong, Khanna, Meng, Liu, Rozi, He, Burke, Lobell, and
  Ermon]{cong2022satmae}
Yezhen Cong, Samar Khanna, Chenlin Meng, Patrick Liu, Erik Rozi, Yutong He,
  Marshall Burke, David~B. Lobell, and Stefano Ermon.
\newblock {SatMAE}: Pre-training transformers for temporal and multi-spectral
  satellite imagery.
\newblock In \emph{Advances in Neural Information Processing Systems
  (NeurIPS)}, pages 197--211, 2022.

\bibitem[Corley et~al.(2024)Corley, Robinson, Dodhia, Lavista~Ferres, and
  Najafirad]{corley2024revisiting}
Isaac Corley, Caleb Robinson, Rahul Dodhia, Juan~M. Lavista~Ferres, and Peyman
  Najafirad.
\newblock Revisiting pre-trained remote sensing model benchmarks: Resizing and
  normalization matters.
\newblock In \emph{IEEE/CVF Conference on Computer Vision and Pattern
  Recognition Workshops (CVPRW)}, pages 3162--3172, 2024.

\bibitem[Darcet et~al.(2024)Darcet, Oquab, Mairal, and
  Bojanowski]{darcet2024registers}
Timoth{\'e}e Darcet, Maxime Oquab, Julien Mairal, and Piotr Bojanowski.
\newblock Vision transformers need registers.
\newblock In \emph{International Conference on Learning Representations
  (ICLR)}, 2024.

\bibitem[Deng et~al.(2009)Deng, Dong, Socher, Li, Li, and
  Fei-Fei]{deng2009imagenet}
Jia Deng, Wei Dong, Richard Socher, Li-Jia Li, Kai Li, and Li Fei-Fei.
\newblock {ImageNet}: A large-scale hierarchical image database.
\newblock In \emph{IEEE Conference on Computer Vision and Pattern Recognition
  (CVPR)}, pages 248--255, 2009.

\bibitem[Dietterich(1998)]{dietterich1998tests}
Thomas~G. Dietterich.
\newblock Approximate statistical tests for comparing supervised classification
  learning algorithms.
\newblock \emph{Neural Computation}, 10\penalty0 (7):\penalty0 1895--1923,
  1998.

\bibitem[Dosovitskiy et~al.(2021)Dosovitskiy, Beyer, Kolesnikov, Weissenborn,
  Zhai, Unterthiner, Dehghani, Minderer, Heigold, Gelly, Uszkoreit, and
  Houlsby]{dosovitskiy2021vit}
Alexey Dosovitskiy, Lucas Beyer, Alexander Kolesnikov, Dirk Weissenborn,
  Xiaohua Zhai, Thomas Unterthiner, Mostafa Dehghani, Matthias Minderer, Georg
  Heigold, Sylvain Gelly, Jakob Uszkoreit, and Neil Houlsby.
\newblock An image is worth 16x16 words: Transformers for image recognition at
  scale.
\newblock In \emph{International Conference on Learning Representations
  (ICLR)}, 2021.

\bibitem[Fuller et~al.(2023)Fuller, Millard, and Green]{fuller2023croma}
Anthony Fuller, Koreen Millard, and James~R. Green.
\newblock {CROMA}: Remote sensing representations with contrastive
  radar-optical masked autoencoders.
\newblock In \emph{Advances in Neural Information Processing Systems
  (NeurIPS)}, pages 5506--5538, 2023.

\bibitem[Grill et~al.(2020)Grill, Strub, Altch{\'e}, Tallec, Richemond,
  Buchatskaya, Doersch, Avila~Pires, Guo, Gheshlaghi~Azar,
  et~al.]{grill2020byol}
Jean-Bastien Grill, Florian Strub, Florent Altch{\'e}, Corentin Tallec,
  Pierre~H. Richemond, Elena Buchatskaya, Carl Doersch, Bernardo Avila~Pires,
  Zhaohan~Daniel Guo, Mohammad Gheshlaghi~Azar, et~al.
\newblock Bootstrap your own latent: A new approach to self-supervised
  learning.
\newblock In \emph{Advances in Neural Information Processing Systems
  (NeurIPS)}, 2020.

\bibitem[He et~al.(2016)He, Zhang, Ren, and Sun]{he2016resnet}
Kaiming He, Xiangyu Zhang, Shaoqing Ren, and Jian Sun.
\newblock Deep residual learning for image recognition.
\newblock In \emph{IEEE Conference on Computer Vision and Pattern Recognition
  (CVPR)}, pages 770--778, 2016.

\bibitem[He et~al.(2020)He, Fan, Wu, Xie, and Girshick]{he2020moco}
Kaiming He, Haoqi Fan, Yuxin Wu, Saining Xie, and Ross Girshick.
\newblock Momentum contrast for unsupervised visual representation learning.
\newblock In \emph{IEEE/CVF Conference on Computer Vision and Pattern
  Recognition (CVPR)}, pages 9726--9735, 2020.

\bibitem[He et~al.(2022)He, Chen, Xie, Li, Doll{\'a}r, and Girshick]{he2022mae}
Kaiming He, Xinlei Chen, Saining Xie, Yanghao Li, Piotr Doll{\'a}r, and Ross
  Girshick.
\newblock Masked autoencoders are scalable vision learners.
\newblock In \emph{IEEE/CVF Conference on Computer Vision and Pattern
  Recognition (CVPR)}, pages 15979--15988, 2022.

\bibitem[Helber et~al.(2019)Helber, Bischke, Dengel, and
  Borth]{helber2019eurosat}
Patrick Helber, Benjamin Bischke, Andreas Dengel, and Damian Borth.
\newblock {EuroSAT}: A novel dataset and deep learning benchmark for land use
  and land cover classification.
\newblock \emph{IEEE Journal of Selected Topics in Applied Earth Observations
  and Remote Sensing}, 12\penalty0 (7):\penalty0 2217--2226, 2019.

\bibitem[Holm(1979)]{holm1979}
Sture Holm.
\newblock A simple sequentially rejective multiple test procedure.
\newblock \emph{Scandinavian Journal of Statistics}, 6\penalty0 (2):\penalty0
  65--70, 1979.

\bibitem[Ilharco et~al.(2021)Ilharco, Wortsman, Carlini, Taori, Dave, Shankar,
  Namkoong, Miller, Hajishirzi, Farhadi, and Schmidt]{ilharco2021openclip}
Gabriel Ilharco, Mitchell Wortsman, Nicholas Carlini, Rohan Taori, Achal Dave,
  Vaishaal Shankar, Hongseok Namkoong, John Miller, Hannaneh Hajishirzi, Ali
  Farhadi, and Ludwig Schmidt.
\newblock {OpenCLIP}.
\newblock Zenodo, \url{https://doi.org/10.5281/zenodo.5143773}, 2021.

\bibitem[Jakubik et~al.(2023)Jakubik, Roy, Phillips, Fraccaro, Godwin,
  et~al.]{jakubik2023prithvi}
Johannes Jakubik, Sujit Roy, C.~E. Phillips, Paolo Fraccaro, Denys Godwin,
  et~al.
\newblock Foundation models for generalist geospatial artificial intelligence.
\newblock \emph{arXiv preprint arXiv:2310.18660}, 2023.

\bibitem[Kolesnikov et~al.(2019)Kolesnikov, Zhai, and
  Beyer]{kolesnikov2019revisiting}
Alexander Kolesnikov, Xiaohua Zhai, and Lucas Beyer.
\newblock Revisiting self-supervised visual representation learning.
\newblock In \emph{IEEE/CVF Conference on Computer Vision and Pattern
  Recognition (CVPR)}, pages 1920--1929, 2019.

\bibitem[Kornblith et~al.(2019)Kornblith, Shlens, and
  Le]{kornblith2019transfer}
Simon Kornblith, Jonathon Shlens, and Quoc~V. Le.
\newblock Do better {ImageNet} models transfer better?
\newblock In \emph{IEEE/CVF Conference on Computer Vision and Pattern
  Recognition (CVPR)}, pages 2656--2666, 2019.

\bibitem[Lacoste et~al.(2023)Lacoste, Lehmann, Rodriguez, Sherwin, Kerner,
  L{\"u}tjens, Irvin, Dao, Alemohammad, Drouin, et~al.]{lacoste2023geobench}
Alexandre Lacoste, Nils Lehmann, Pau Rodriguez, Evan~David Sherwin, Hannah
  Kerner, Bj{\"o}rn L{\"u}tjens, Jeremy~Andrew Irvin, David Dao, Hamed
  Alemohammad, Alexandre Drouin, et~al.
\newblock {GEO-Bench}: Toward foundation models for earth monitoring.
\newblock In \emph{Advances in Neural Information Processing Systems (NeurIPS),
  Datasets and Benchmarks Track}, pages 51080--51093, 2023.

\bibitem[Liu and Nocedal(1989)]{liu1989lbfgs}
Dong~C. Liu and Jorge Nocedal.
\newblock On the limited memory {BFGS} method for large scale optimization.
\newblock \emph{Mathematical Programming}, 45\penalty0 (1--3):\penalty0
  503--528, 1989.

\bibitem[Liu et~al.(2024)Liu, Chen, Guan, Zhou, Zhu, Ye, Fu, and
  Zhou]{liu2024remoteclip}
Fan Liu, Delong Chen, Zhangqingyun Guan, Xiaocong Zhou, Jiale Zhu, Qiaolin Ye,
  Liyong Fu, and Jun Zhou.
\newblock {RemoteCLIP}: A vision language foundation model for remote sensing.
\newblock \emph{IEEE Transactions on Geoscience and Remote Sensing},
  62:\penalty0 1--16, 2024.

\bibitem[Loshchilov and Hutter(2019)]{loshchilov2019adamw}
Ilya Loshchilov and Frank Hutter.
\newblock Decoupled weight decay regularization.
\newblock In \emph{International Conference on Learning Representations
  (ICLR)}, 2019.

\bibitem[Lu et~al.(2018)Lu, Wang, Zheng, and Li]{lu2018rsicd}
Xiaoqiang Lu, Binqiang Wang, Xiangtao Zheng, and Xuelong Li.
\newblock Exploring models and data for remote sensing image caption
  generation.
\newblock \emph{IEEE Transactions on Geoscience and Remote Sensing},
  56\penalty0 (4):\penalty0 2183--2195, 2018.

\bibitem[Ma{\~n}as et~al.(2021)Ma{\~n}as, Lacoste, Gir{\'o}-i Nieto, Vazquez,
  and Rodr{\'i}guez]{manas2021seco}
Oscar Ma{\~n}as, Alexandre Lacoste, Xavier Gir{\'o}-i Nieto, David Vazquez, and
  Pau Rodr{\'i}guez.
\newblock Seasonal contrast: Unsupervised pre-training from uncurated remote
  sensing data.
\newblock In \emph{IEEE/CVF International Conference on Computer Vision
  (ICCV)}, pages 9394--9403, 2021.

\bibitem[Marsocci et~al.(2026)Marsocci, Jia, Le~Bellier, Kerekes, Zeng,
  et~al.]{marsocci2024pangaea}
Valerio Marsocci, Yuru Jia, Georges Le~Bellier, David Kerekes, Liang Zeng,
  et~al.
\newblock {PANGAEA}: Assessing geospatial foundation models capabilities
  through a global and inclusive benchmark.
\newblock \emph{IEEE Geoscience and Remote Sensing Magazine}, 14\penalty0
  (1):\penalty0 245--285, 2026.

\bibitem[McNemar(1947)]{mcnemar1947}
Quinn McNemar.
\newblock Note on the sampling error of the difference between correlated
  proportions or percentages.
\newblock \emph{Psychometrika}, 12\penalty0 (2):\penalty0 153--157, 1947.

\bibitem[Nadeau and Bengio(2003)]{nadeau2003inference}
Claude Nadeau and Yoshua Bengio.
\newblock Inference for the generalization error.
\newblock \emph{Machine Learning}, 52\penalty0 (3):\penalty0 239--281, 2003.

\bibitem[Neumann et~al.(2019)Neumann, Pinto, Zhai, and
  Houlsby]{neumann2019indomain}
Maxim Neumann, Andre~Susano Pinto, Xiaohua Zhai, and Neil Houlsby.
\newblock In-domain representation learning for remote sensing.
\newblock \emph{arXiv preprint arXiv:1911.06721}, 2019.

\bibitem[Oquab et~al.(2024)Oquab, Darcet, Moutakanni, Vo, Szafraniec, Khalidov,
  Fernandez, Haziza, Massa, El-Nouby, et~al.]{oquab2024dinov2}
Maxime Oquab, Timoth{\'e}e Darcet, Th{\'e}o Moutakanni, Huy Vo, Marc
  Szafraniec, Vasil Khalidov, Pierre Fernandez, Daniel Haziza, Francisco Massa,
  Alaaeldin El-Nouby, et~al.
\newblock {DINOv2}: Learning robust visual features without supervision.
\newblock \emph{Transactions on Machine Learning Research (TMLR)}, 2024.

\bibitem[Paszke et~al.(2019)Paszke, Gross, Massa, Lerer, Bradbury,
  et~al.]{paszke2019pytorch}
Adam Paszke, Sam Gross, Francisco Massa, Adam Lerer, James Bradbury, et~al.
\newblock {PyTorch}: An imperative style, high-performance deep learning
  library.
\newblock In \emph{Advances in Neural Information Processing Systems
  (NeurIPS)}, 2019.

\bibitem[Pedregosa et~al.(2011)Pedregosa, Varoquaux, Gramfort, Michel, Thirion,
  et~al.]{pedregosa2011sklearn}
Fabian Pedregosa, Ga{\"e}l Varoquaux, Alexandre Gramfort, Vincent Michel,
  Bertrand Thirion, et~al.
\newblock Scikit-learn: Machine learning in {P}ython.
\newblock \emph{Journal of Machine Learning Research}, 12:\penalty0 2825--2830,
  2011.

\bibitem[Psomas et~al.(2026)Psomas, Christopoulos, Baltzi, Kakogeorgiou,
  Aravanis, Komodakis, Karantzalos, Avrithis, and Tolias]{psomas2026ep}
Bill Psomas, Dionysis Christopoulos, Eirini Baltzi, Ioannis Kakogeorgiou,
  Tilemachos Aravanis, Nikos Komodakis, Konstantinos Karantzalos, Yannis
  Avrithis, and Giorgos Tolias.
\newblock Attention, please! {R}evisiting attentive probing through the lens of
  efficiency.
\newblock In \emph{International Conference on Learning Representations
  (ICLR)}, 2026.

\bibitem[Qu et~al.(2016)Qu, Li, Tao, and Lu]{qu2016ucmcaptions}
Bo Qu, Xuelong Li, Dacheng Tao, and Xiaoqiang Lu.
\newblock Deep semantic understanding of high resolution remote sensing image.
\newblock In \emph{International Conference on Computer, Information and
  Telecommunication Systems (CITS)}, pages 1--5, 2016.

\bibitem[Radford et~al.(2021)Radford, Kim, Hallacy, Ramesh, Goh, Agarwal,
  Sastry, Askell, Mishkin, Clark, Krueger, and Sutskever]{radford2021clip}
Alec Radford, Jong~Wook Kim, Chris Hallacy, Aditya Ramesh, Gabriel Goh,
  Sandhini Agarwal, Girish Sastry, Amanda Askell, Pamela Mishkin, Jack Clark,
  Gretchen Krueger, and Ilya Sutskever.
\newblock Learning transferable visual models from natural language
  supervision.
\newblock In \emph{International Conference on Machine Learning (ICML)}, pages
  8748--8763, 2021.

\bibitem[Reed et~al.(2023)Reed, Gupta, Li, Brockman, Funk, Clipp, Keutzer,
  Candido, Uyttendaele, and Darrell]{reed2023scalemae}
Colorado~J. Reed, Ritwik Gupta, Shufan Li, Sarah Brockman, Christopher Funk,
  Brian Clipp, Kurt Keutzer, Salvatore Candido, Matt Uyttendaele, and Trevor
  Darrell.
\newblock {Scale-MAE}: A scale-aware masked autoencoder for multiscale
  geospatial representation learning.
\newblock In \emph{IEEE/CVF International Conference on Computer Vision
  (ICCV)}, pages 4065--4076, 2023.

\bibitem[Risojevi{\'c} and Stojni{\'c}(2022)]{risojevic2021imagenet}
Vladimir Risojevi{\'c} and Vladan Stojni{\'c}.
\newblock Do we still need {ImageNet} pre-training in remote sensing scene
  classification?
\newblock \emph{The International Archives of the Photogrammetry, Remote
  Sensing and Spatial Information Sciences}, XLIII-B3-2022:\penalty0
  1399--1406, 2022.

\bibitem[Rousseeuw(1987)]{rousseeuw1987silhouettes}
Peter~J. Rousseeuw.
\newblock Silhouettes: A graphical aid to the interpretation and validation of
  cluster analysis.
\newblock \emph{Journal of Computational and Applied Mathematics}, 20:\penalty0
  53--65, 1987.

\bibitem[Sim{\'e}oni et~al.(2025)Sim{\'e}oni, Vo, Seitzer, Baldassarre, Oquab,
  Jose, Khalidov, Szafraniec, Yi, Ramamonjisoa, et~al.]{simeoni2025dinov3}
Oriane Sim{\'e}oni, Huy~V. Vo, Maximilian Seitzer, Federico Baldassarre, Maxime
  Oquab, Cijo Jose, Vasil Khalidov, Marc Szafraniec, Seungeun Yi, Micha{\"e}l
  Ramamonjisoa, et~al.
\newblock {DINOv3}.
\newblock \emph{arXiv preprint arXiv:2508.10104}, 2025.

\bibitem[Stewart et~al.(2022)Stewart, Robinson, Corley, Ortiz, Lavista~Ferres,
  and Banerjee]{stewart2022torchgeo}
Adam~J. Stewart, Caleb Robinson, Isaac~A. Corley, Anthony Ortiz, Juan~M.
  Lavista~Ferres, and Arindam Banerjee.
\newblock {TorchGeo}: Deep learning with geospatial data.
\newblock In \emph{ACM SIGSPATIAL International Conference on Advances in
  Geographic Information Systems}, pages 1--12, 2022.

\bibitem[Straka and Gruber(2025)]{straka2025satdino}
Jakub Straka and Ivan Gruber.
\newblock {SatDINO}: A deep dive into self-supervised pretraining for remote
  sensing.
\newblock \emph{arXiv preprint arXiv:2508.21402}, 2025.

\bibitem[van~der Maaten and Hinton(2008)]{vandermaaten2008tsne}
Laurens van~der Maaten and Geoffrey Hinton.
\newblock Visualizing data using t-{SNE}.
\newblock \emph{Journal of Machine Learning Research}, 9:\penalty0 2579--2605,
  2008.

\bibitem[Wang et~al.(2022)Wang, Albrecht, Braham, Mou, and
  Zhu]{wang2022sslreview}
Yi Wang, Conrad~M. Albrecht, Nassim Ait~Ali Braham, Lichao Mou, and Xiao~Xiang
  Zhu.
\newblock Self-supervised learning in remote sensing: A review.
\newblock \emph{IEEE Geoscience and Remote Sensing Magazine}, 10\penalty0
  (4):\penalty0 213--247, 2022.

\bibitem[Wang et~al.(2023)Wang, Braham, Xiong, Liu, Albrecht, and
  Zhu]{wang2023ssl4eo}
Yi Wang, Nassim Ait~Ali Braham, Zhitong Xiong, Chenying Liu, Conrad~M.
  Albrecht, and Xiao~Xiang Zhu.
\newblock {SSL4EO-S12}: A large-scale multimodal, multitemporal dataset for
  self-supervised learning in {Earth} observation.
\newblock \emph{IEEE Geoscience and Remote Sensing Magazine}, 11\penalty0
  (3):\penalty0 98--106, 2023.

\bibitem[Wightman(2019)]{wightman2019timm}
Ross Wightman.
\newblock {PyTorch} image models.
\newblock \url{https://github.com/huggingface/pytorch-image-models}, 2019.

\bibitem[Wilcoxon(1945)]{wilcoxon1945}
Frank Wilcoxon.
\newblock Individual comparisons by ranking methods.
\newblock \emph{Biometrics Bulletin}, 1\penalty0 (6):\penalty0 80--83, 1945.

\bibitem[Wolf et~al.(2020)Wolf, Debut, Sanh, Chaumond, Delangue,
  et~al.]{wolf2020transformers}
Thomas Wolf, Lysandre Debut, Victor Sanh, Julien Chaumond, Clement Delangue,
  et~al.
\newblock Transformers: State-of-the-art natural language processing.
\newblock In \emph{Conference on Empirical Methods in Natural Language
  Processing: System Demonstrations (EMNLP)}, pages 38--45, 2020.

\bibitem[Xiong et~al.(2024)Xiong, Wang, Zhang, Stewart, Hanna, Borth,
  Papoutsis, Le~Saux, Camps-Valls, and Zhu]{xiong2024dofa}
Zhitong Xiong, Yi Wang, Fahong Zhang, Adam~J. Stewart, Jo{\"e}lle Hanna, Damian
  Borth, Ioannis Papoutsis, Bertrand Le~Saux, Gustau Camps-Valls, and
  Xiao~Xiang Zhu.
\newblock Neural plasticity-inspired multimodal foundation model for {Earth}
  observation.
\newblock \emph{arXiv preprint arXiv:2403.15356}, 2024.

\bibitem[Yang and Newsam(2010)]{yang2010ucm}
Yi Yang and Shawn Newsam.
\newblock Bag-of-visual-words and spatial extensions for land-use
  classification.
\newblock In \emph{ACM SIGSPATIAL International Conference on Advances in
  Geographic Information Systems}, pages 270--279, 2010.

\bibitem[Yuan et~al.(2022)Yuan, Zhang, Fu, Li, Deng, Wang, and
  Sun]{yuan2022rsitmd}
Zhiqiang Yuan, Wenkai Zhang, Kun Fu, Xuan Li, Chubo Deng, Hongqi Wang, and Xian
  Sun.
\newblock Exploring a fine-grained multiscale method for cross-modal remote
  sensing image retrieval.
\newblock \emph{IEEE Transactions on Geoscience and Remote Sensing},
  60:\penalty0 1--19, 2022.

\bibitem[Zhou et~al.(2022)Zhou, Wei, Wang, Shen, Xie, Yuille, and
  Kong]{zhou2022ibot}
Jinghao Zhou, Chen Wei, Huiyu Wang, Wei Shen, Cihang Xie, Alan Yuille, and Tao
  Kong.
\newblock {iBOT}: Image {BERT} pre-training with online tokenizer.
\newblock In \emph{International Conference on Learning Representations
  (ICLR)}, 2022.

\bibitem[Zhou et~al.(2018)Zhou, Newsam, Li, and Shao]{zhou2018patternnet}
Weixun Zhou, Shawn Newsam, Congmin Li, and Zhenfeng Shao.
\newblock {PatternNet}: A benchmark dataset for performance evaluation of
  remote sensing image retrieval.
\newblock \emph{ISPRS Journal of Photogrammetry and Remote Sensing},
  145:\penalty0 197--209, 2018.

\end{thebibliography}
